\documentclass[runningheads]{llncs}

\usepackage{eccv}

\usepackage{eccvabbrv}

\usepackage{graphicx}
\usepackage{booktabs}

\usepackage[accsupp]{axessibility}  

\usepackage[pagebackref,breaklinks,colorlinks,citecolor=eccvblue]{hyperref}

\usepackage{orcidlink}

\usepackage{bm}
\usepackage{multirow}
\usepackage{array}

\usepackage{hyperref}
\usepackage{xcolor}
\usepackage{booktabs}
\hypersetup{
    colorlinks=true,
    citecolor=green,
    linkcolor=red,
    urlcolor=green,
}

\newcolumntype{C}[1]{>{\centering\arraybackslash}p{#1}}
\newcolumntype{R}[1]{>{\raggedleft\arraybackslash}p{#1}}
\newcolumntype{L}[1]{>{\raggedright\arraybackslash}p{#1}}

\begin{document}

\title{TAG: Guidance-free Open-Vocabulary \\Semantic Segmentation} 

\titlerunning{Abbreviated paper title}

\author{Yasufumi Kawano\inst{1}\orcidlink{0000-0002-2252-1910} \and
Yoshimitsu Aoki\inst{1}\orcidlink{0000-0001-7361-0027}}

\authorrunning{Y.~Kawano and Y.~Aoki.}

\institute{Keio University, Japan \\
\email{ykawano@aoki-medialab.jp, aoki@elec.keio.ac.jp}}

\maketitle

\begin{abstract}
Semantic segmentation is a crucial task in computer vision, where each pixel in an image is classified into a category. However, traditional methods face significant challenges, including the need for pixel-level annotations and extensive training. Furthermore, because supervised learning uses a limited set of predefined categories, models typically struggle with rare classes and cannot recognize new ones. Unsupervised and open-vocabulary segmentation, proposed to tackle these issues, faces challenges, including the inability to assign specific class labels to clusters and the necessity of user-provided text queries for guidance. In this context, we propose a novel approach, \textbf{TAG} which achieves \textbf{T}raining, \textbf{A}nnotation, and \textbf{G}uidance-free open-vocabulary semantic segmentation. TAG utilizes pre-trained models such as CLIP and DINO to segment images into meaningful categories without additional training or dense annotations. It retrieves class labels from an external database, providing flexibility to adapt to new scenarios. Our TAG achieves state-of-the-art results on PascalVOC, PascalContext and ADE20K for open-vocabulary segmentation without given class names, i.e. improvement of +15.3 mIoU on PascalVOC. All code and data will be released at \url{https://github.com/Valkyrja3607/TAG}.

\end{abstract}

\begin{figure*}[t]
    \centering
    \begin{minipage}[b]{\linewidth}
    \includegraphics[width=\linewidth]{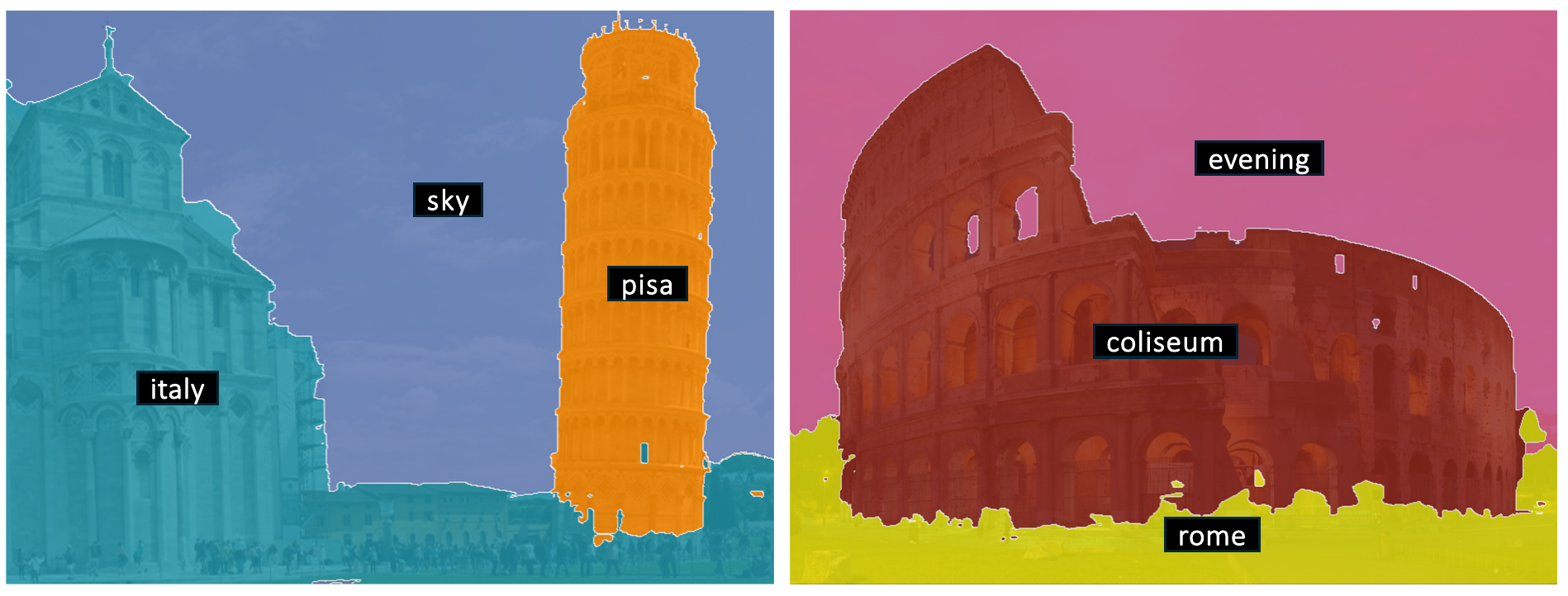}
    \caption{\textbf{Guidance-free Open-Vocabulary Semantic Segmentation.} Our TAG can segment an image into meaningful segments without training, annotation, or guidance. It successfully segments structures such as the Leaning Tower of Pisa and the Colosseum. Unlike traditional open-vocabulary semantic segmentation methods, TAG can segment and categorize without text-guidance.}
    \label{fig:demo}
    \end{minipage}
\end{figure*}

\section{Introduction}
\label{sec:intro}
Semantic segmentation represents a crucial task in computer vision, which describes assigning class labels to each pixel of an image. Its applications span diverse domains, including robotics and satellite image analysis. 

Despite its significance, current semantic segmentation methods still face several critical challenges.
Firstly, these methods are high-cost, requiring pixel-level annotation and extensive training. Secondly, since supervised learning depends on a predefined set of categories, detecting extremely rare or completely new classes during prediction is virtually impossible.

Two related tasks were proposed to address these limitations: unsupervised and open-vocabulary semantic segmentation.
Unsupervised semantic segmentation~\cite{picie,stego,hp} avoids the expensive annotation process by using representations obtained through a backbone model~\cite{dino,dinov2} trained on a different task. 
Open-vocabulary semantic segmentation~\cite{sam, openseg, odise, maskclip, openseg2} enables the identification of a wide array of categories through natural language and is not bound to a pre-defined set of categories.

However, there are still challenges to be solved with these methods.
Unsupervised semantic segmentation clusters images by class but cannot identify the class of each cluster, while open-vocabulary segmentation assumes that text queries describing objects in the image are provided by the user.
To address these challenges, zero-guidance segmentation emerged in \cite{zeroseg}, enabling open-vocabulary segmentation without the need for inputting class candidates (guidance), yet there is still room for improvement in terms of performance.
We categorize these related works into four distinct areas in Table~\ref{tab:relationship}.

Based on these backgrounds, we further improved this approach by introducing a novel method named \textbf{TAG}, which offers higher performance and flexibility. As its primary strength, TAG achieves \textbf{T}raining, \textbf{A}nnotation, and \textbf{G}uidance-free open-vocabulary semantic segmentation. This method employs a novel approach by extracting semantic features from each pixel in an image using CLIP~\cite{clip}, and then retrieving the open-vocabulary classes based on these features from an external database~\cite{pmd, cc12m, english, wordnet}.
TAG operates using pre-trained frozen models CLIP~\cite{clip} and DINOv2~\cite{dinov2}, eliminating the need for an additional training process. These models do not utilize the dense and costly annotations traditionally required for semantic segmentation. Furthermore, through the extensibility of its database, this method also incorporates flexibility, making it easy to adapt to new classes or scenarios.
A major distinction between previous methods~\cite{zeroseg, selfseg}, and our TAG is that it provides more flexibility as it can be extended to include new concepts by adding them to the database while previous methods require re-training.
It is important to note that while the database used in TAG is finite, the language models like BLIP~\cite{blip} or GPT~\cite{gpt2} are also constructed from similarly finite datasets. In \cite{vic}, it is even reported the retrieval-based methods provided superior results over BLIP~\cite{blip} in the context of image classification.


Our TAG can segment an image into meaningful segments as shown in Figure~\ref{fig:demo} without any text guidance. In particular, TAG is able to accurately segment structures with their proper nouns, such as the \textit{Leaning Tower of Pisa} and the \textit{Coliseum}.
In addition, TAG shows significant improvements in contrast to other comparable segmentation methods, i.e. on the PascalVOC~\cite{voc} dataset (+15.3 mIoU).

Our contributions are the following:
\begin{enumerate}
    \item We propose a novel approach, namely TAG, to achieve open-vocabulary semantic segmentation that does not require pre-defined categories by retrieving segment categories from an external database.
    \item TAG achieves compelling segmentation results for all categories in the wild without any additional training, high-cost dense annotation, or text query guidance.
    \item TAG outperforms the previous state-of-the-art methods by 15.3 mIoU on the PascalVOC~\cite{voc} dataset, demonstrating the superior segmentation performance of our proposed approach.
\end{enumerate}

\begin{table}[t]
  \caption{\textbf{Relationship with related works.} We categorize related works into four distinct areas: training-free, (dense) annotation-free, guidance-free, and open-vocabulary.}
  \label{tab:relationship}
  \centering
  \scalebox{0.9}[0.9]{
  \begin{tabular}{l | c | c | c | c | c }
  \hline
    \multirow{2}{*}{Method} & Pretrained & Training & Annotation & Guidance & Open \\
     &  Backbone & Free & Free & Free & Vocabulary \\
     \hline
     \multicolumn{6}{l}{\textbf{Traditional Semantic Segmentation}} \\
     \hline
     DeepLab~\cite{deeplab} & - & - & - & - & - \\
     \hline
     \multicolumn{6}{l}{\textbf{Unsupervised Segmentation}} \\
     \hline
     STEGO~\cite{stego} & DINO~\cite{dino} & - & $\surd$ & $\surd$ & - \\
     HP~\cite{hp} & DINOv2~\cite{dinov2} & - & $\surd$ & $\surd$ & - \\
     \hline
     \multicolumn{6}{l}{\textbf{Open Vocabulary Segmentation}} \\
     \hline
     ODISE~\cite{odise} & StableDiffusion~\cite{sd} & - & - & - & $\surd$ \\
     MaskCLIP~\cite{maskclip} & CLIP~\cite{clip} & $\surd$ & $\surd$ & - & $\surd$ \\
     \hline
     \multicolumn{6}{l}{\textbf{Zero-Guidance Segmentation}} \\
     \hline
     ZeroSeg~\cite{zeroseg} & CLIP\cite{clip}\&~DINO\cite{dino}\&~GPT-2\cite{gpt2} & $\surd$ & $\surd$ 
     & $\surd$ & $\surd$ \\
     SelfSeg~\cite{selfseg} & BLIP\cite{blip}~\&~X-Decoder\cite{xd} & $\surd$ & - & $\surd$ & $\surd$ \\
     TAG (Ours) & CLIP\cite{clip}~\&~DINOv2\cite{dinov2} & $\surd$ & $\surd$ & $\surd$ & $\surd$ \\
    \hline
  \end{tabular}
  }
\end{table}

\section{Related Work}
\subsection{Semantic Segmentation}
Semantic segmentation is the task of assigning class labels to all pixels in an image, commonly using convolutional neural networks~\cite{deeplab, long} or vision transformers~\cite{transformer} for end-to-end training. These methods, while effective, depend on extensive annotation and significant computational resources for training, and are limited to predefined categories. Thus, unsupervised, and domain-flexible approaches have recently gained importance.

Unsupervised semantic segmentation~\cite{iic,picie,stego,hp} attempts to solve semantic segmentation without using any kind of supervision.
STEGO~\cite{stego} and HP~\cite{hp} optimize the head of a segmentation model using image features obtained from a backbone pre-trained by DINO~\cite{dino} and DINOv2~\cite{dinov2}, an unsupervised method for many tasks.
However, unsupervised semantic segmentation clusters images by class but cannot identify the class of the each cluster.
In contrast, our TAG distinguishes classes without extra training or annotation.

\subsection{Open-Vocabulary Semantic Segmentation}
Open vocabulary semantic segmentation, crucial for segmenting objects across domains without being limited to predefined categories, has seen notable advancements with the introduction of key methodologies~\cite{sam,maskclip,clipseg,reco, zeroshot,seem, spnet, lseg, openseg, openseg2, odise,opsnet, ovseg}.

Early attempts, such as ZS3Net~\cite{zeroshot} and SPNet~\cite{spnet}, focused on zero-shot learning, training custom modules to bridge visual and language embedding spaces. These methods set the foundation for future improvements.

This area has seen significant improvement, particularly through integrating vision-language models like CLIP~\cite{clip}, which train visual and textual feature encoders on extensive image-text pairs.
LSeg~\cite{lseg}, OpenSeg~\cite{openseg}, OPSNet~\cite{opsnet}, and OVSeg~\cite{ovseg} have each contributed to the advancements in the field leveraging CLIP~\cite{clip}. These methods typically generate class-agnostic masks before using CLIP~\cite{clip} to classify each mask, demonstrating the versatility of CLIP~\cite{clip} embeddings in open vocabulary semantic segmentation.

Moreover, MaskCLIP~\cite{maskclip} and GEM~\cite{gem} have highlighted the potential of using intermediate representations from a frozen CLIP~\cite{clip} encoder to directly segment images without additional training, reducing both annotation and training costs.
Concurrently, models like ODISE~\cite{odise} have explored the integration of pre-trained diffusion models~\cite{sd} with CLIP~\cite{clip} to extend to high performance panoptic segmentation.

Despite these advancements, a limitation across these methods is their reliance on text input as guidance from users. 
Our TAG tackles this limitation and allows for open-vocabulary segmentation without text guidance.
Closest and concurrent to our work is the zero-guidance semantic segmentation paradigm~\cite{zeroseg}, in which clustered DINO~\cite{dino} embeddings are combined with CLIP~\cite{clip}. To generate captions from CLIP~\cite{clip} features, ZeroSeg~\cite{zeroseg} uses ZeroCap~\cite{zerocap} which combines a language model, GPT-2~\cite{gpt2}, with CLIP~\cite{clip}. It adjusts parts of GPT-2~\cite{gpt2} to finish the sentence, starting with "Image of a ..." so that the sentence closely matches the images according to CLIP's understanding.

However, there is still room for improvement in terms of performance.
We hypothesize that the issue is related to the performance of ZeroCap~\cite{zerocap}. Therefore, as a new method, our TAG uses a novel approach that retrieves categories from a database for estimating categories.



\subsection{Text Retrieval from CLIP Embedding}
In natural language processing, retrieving information from external databases has been shown to boost the performance of large language models~\cite{retrieval1, retrieval2, retrieval3}. This concept is also explored in computer vision, particularly for addressing class imbalance by using databases to retrieve training samples or image-text pairs. 
RAC~\cite{rac} and VIC~\cite{vic} achieved image classification without relying on predefined classes by utilizing an external database. It has the advantage of low memory consumption because it only uses captions from databases like Public Multimodal Datasets (PMD)~\cite{pmd} collecting image-text pairs from different public datasets.

\begin{figure*}[t]
    \centering
    \begin{minipage}[b]{\linewidth}
    \includegraphics[width=\linewidth]{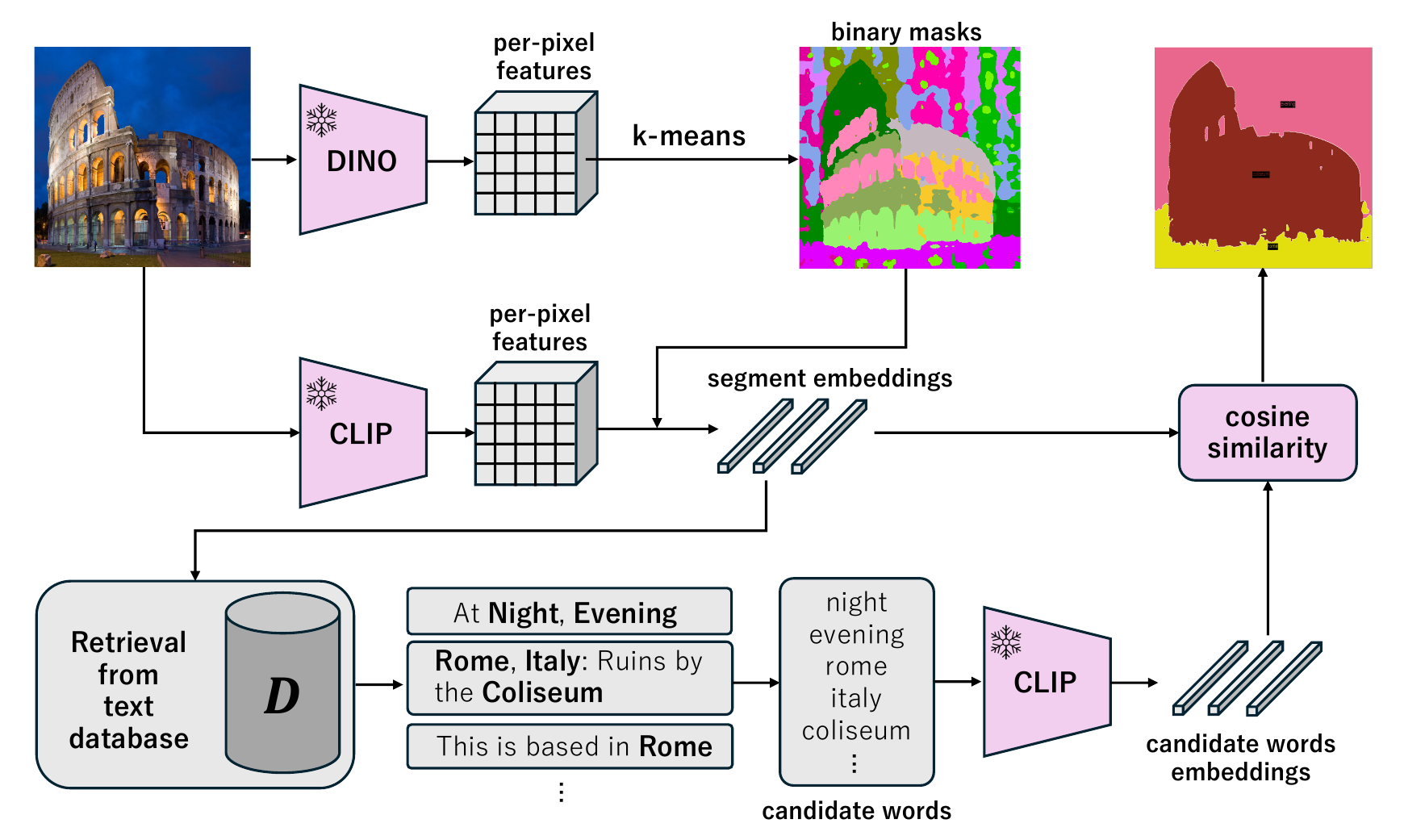}
    \caption{\textbf{High-level overview of our TAG architecture.} Our TAG can partition images into semantic segments and label each segment with open-vocabulary categories. First, TAG identifies segment candidates using per-pixel features obtained from DINOv2~\cite{dinov2}. Then, it acquires representative segment embeddings for segment candidates using per-pixel features from a ViT pre-trained with CLIP~\cite{clip}. Finally, the categories are assigned to each candidate segment by retrieving the closest matching sentence from an external database. Note that the input is only the image, with no need to input category candidates as guidance.}
    \label{fig:overview}
    \end{minipage}
\end{figure*}

\begin{figure*}[t]
    \centering
    \begin{minipage}[b]{\linewidth}
    \includegraphics[width=\linewidth]{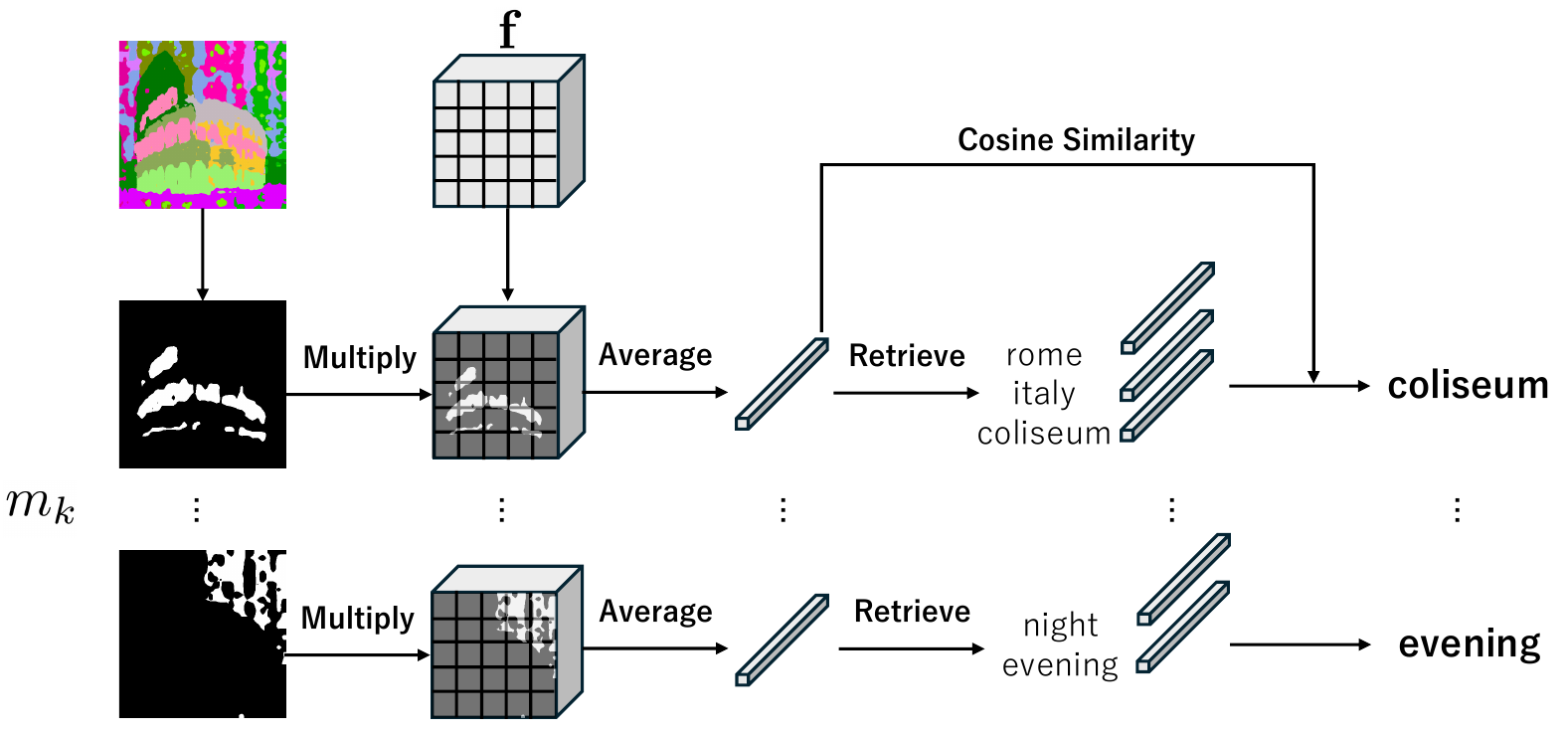}
    \caption{\textbf{Overview of the flow for each segment.} Each segment independently retrieves for category candidates and assigns a category.}
    \label{fig:detail}
    \end{minipage}
\end{figure*}

\section{Method}
Figure~\ref{fig:overview} shows an overview of our proposed method which we call TAG, a novel approach.
Our TAG attempts to partition input images into semantic segments and label each segment with open-vocabulary categories. To this end, we propose to identify segment candidates using per-pixel features obtained from DINOv2~\cite{dinov2} (Sec.~\ref{sec:dino}), acquire representative segment embeddings for segment candidates using per-pixel features from a ViT pre-trained with CLIP~\cite{clip} (Sec.~\ref{sec:clip}), and assign categories to each candidate segment by retrieving the closest matching sentence from an external database (Sec.~\ref{sec:retrieval}).
Note that, unlike traditional open-vocabulary semantic segmentation, the input is only the image, with no need to input category candidates as guidance.

\subsection{Segment Candidates with DINO}
\label{sec:dino}
It has been observed that segmentation results obtained from CLIP-based segmentation methods~\cite{maskclip, gem} are fragmented and noisy as shown in Figure~\ref{fig:clipbase}. Therefore, the first step in our TAG pipeline is calculating segmentation candidates to achieve more accurate segmentation results. To obtain more precise segmentation outcomes than CLIP-based methods without using dense annotations, we reference unsupervised segmentation methods~\cite{stego,hp} and employ a ViT pre-trained with DINOv2~\cite{dinov2}.

The output of DINOv2~\cite{dinov2} is a feature map $\in \mathbb{R}^{D \times \frac{H}{P} \times  \frac{W}{P}}$, where D is the dimension of the feature, $P$ is the patch size of the transformer, and $H$ and $W$ are image sizes.
This feature map will be upsampled to per-pixel features $\in \mathbb{R}^{D \times H \times W}$.
Once the per-pixel feature is obtained, to assign categories for each segmentation candidate, we use k-means to divide the per-pixel features into segmentation candidates, resulting in oversegmentation.

\subsection{Representative Segment Embeddings with CLIP}
\label{sec:clip}
CLIP~\cite{clip} is a ViT model that can embed images and text into the same latent space. To assign natural language categories to each segment, we use CLIP~\cite{clip} to embed the image at the pixel level.

Instead of directly acquiring pixel-level features from CLIP~\cite{clip}, we extract dense patch-level features from the image encoder of CLIP~\cite{clip} following CLIP-based segmentation methods~\cite{maskclip, gem}.
The image encoder of CLIP~\cite{clip} uses a multi-head attention layer, where the globally average-pooled feature works as the query, and the feature at each patch generates a key-value pair. 
Then, this layer outputs a spatial weighted sum of the incoming feature map followed by a linear layer $F(\cdot)$:

\begin{align}
\label{eq:clip}
\text{AttnPool}(\overline{q}, k, v) &= F\left(\sum_{i} \text{softmax}\left(\frac{\overline{q} k_i^{T}}{C}\right) v_i\right) \nonumber\\
&= \sum_{i} \text{softmax}\left(\frac{\overline{q} k_i^{T}}{C}\right) F(v_i),
\end{align}

\begin{equation}
    \overline{q} = \text{Emb}_{q}(\overline{x}), k_i = \text{Emb}_{k}(x_i),v_i = \text{Emb}_{v}(x_i) ,
\end{equation}

\noindent
where $C$ denotes a constant scaling factor and $\text{Emb}(\cdot)$ represents a linear embedding layer. $x_i$ is the input feature at patch $i$ and $\overline{x}$ is the average of all $x_i$.
The Transformer layer in CLIP~\cite{clip} outputs a detailed image representation, made possible because $F(v_i)$, computed at each spatial location, captures a rich response of local semantics.

Based on this observation, we utilize the features from the last attention layer of CLIP~\cite{clip} image encoder by adopting the GEM~\cite{gem} mechanism.

CLIP model in TAG outputs value features $\in \mathbb{R}^{D \times \frac{H}{P} \times  \frac{W}{P}}$, where D is the dimension of the feature, $P$ is the patch size of the transformer. These features contain dense representations of the image, capturing patch-level information, which we upsample to per-pixel features $\in \mathbb{R}^{D \times H \times W}$, corresponding to the same size as the features obtained from DINO~\cite{dinov2}.

Next, to assign categories to segment candidates, we calculate embedding features representing the segments from CLIP~\cite{clip} per-pixel features $\mathbf{f} \in \mathbb{R}^{D \times H \times W}$ as shown in Figure~\ref{fig:detail}.
For each segment $k$, this representative segment embedding $\bar{\mathbf{f}}_k \in \mathbb{R}^{D}$ is computed by averaging based on the values $m_{khw} \in \{0, 1\}$, which results from applying k-means to the output of the DINO~\cite{dinov2} with $k$ classes, as follows:

\begin{align}
\label{eq:ave}
\bar{\mathbf{f}}_{k} &= \frac{1}{M_k} \sum_{h,w} m_{khw} \cdot \mathbf{f}_{hw}, \quad M_k=\sum_{h,w} m_{khw}
\end{align}

\subsection{Segment Category Retrieval}
\label{sec:retrieval}
CLIP~\cite{clip} can embed images and text in the same latent space, but the model itself cannot generate images or text from the embedded features.
To address this challenge, our proposed method TAG finds the closest category using multi-modal data from large databases.

First, we retrieve a few of the most probable candidate classes from the large classification space.

Let $D$ be the database of image captions. Given representative segment embedding $\bar{\mathbf{f}}_k$, retrieve the set $D_{\bar{\mathbf{f}}_k} \subset D$ of the $n$ closest captions to each segment embedding by

\begin{align}
D_{\bar{\mathbf{f}}_k} = \underset{\mathbf{d} \in D}{\text{top-}n} \: \frac{\bar{\mathbf{f}}_k^T \cdot \mathbf{f}_d}{\| \bar{\mathbf{f}}_k \| \cdot  \|\mathbf{f}_d \|}, \quad \mathbf{f}_d = \text{CLIP}_t(d) ,
\end{align}

\noindent
where $\text{CLIP}_t$ is the text encoder of the CLIP~\cite{clip}.

Next, to extract candidate words $C_{\bar{\mathbf{f}}_k}$ from the set $D_{\bar{\mathbf{f}}_k}$, we create a set of all words that are contained in the captions. 
We sequentially apply three operations: (i) remove noisy candidates, (ii) standardize their format, and (iii) filter them.

In the first operation, we remove all the irrelevant words, such as URLs, or file extensions.

Secondly, we align words referring to the same semantic class in a standardized format.
Specifically, Converting upper case to lower case and plural words to singular format.

In the final operation, we filter out two types of words: rare and noisy categories based on the frequency of word occurrences, as well as entire categories of words determined by Part-Of-Speech (POS)~\cite{pos} tagging.
Frequency filtering involves retaining only those words that appear more than two times in the input text. If the threshold is set too high and no words meet the criterion, it is lowered to include at least the most frequently occurring words.
The POS~\cite{pos} tagging classifies words into groups like adjectives, articles, nouns, or verbs, allowing us to exclude any terms that do not hold semantic significance as segmentation categories.

Given candidate words $C_{\bar{\mathbf{f}}_k}$, we assign words to representative segment embedding $\bar{\mathbf{f}}_k$ as

\begin{align}
W = \underset{\mathbf{c} \in C_{\bar{\mathbf{f}}_k^T}}{\text{argmax}} \: \frac{\bar{\mathbf{f}}_k^T \cdot \mathbf{f}_c}{\| \bar{\mathbf{f}}_k \| \cdot  \|\mathbf{f}_c \|}, \quad \mathbf{f}_c = \text{CLIP}_t(c) ,
\end{align}

\noindent
where W is assigned a category word to segment.
Through the above process, we can obtain segmentation results by assigning categories to each segment candidate.

\begin{figure*}[t]
    \centering
    \includegraphics[width=0.8\linewidth]{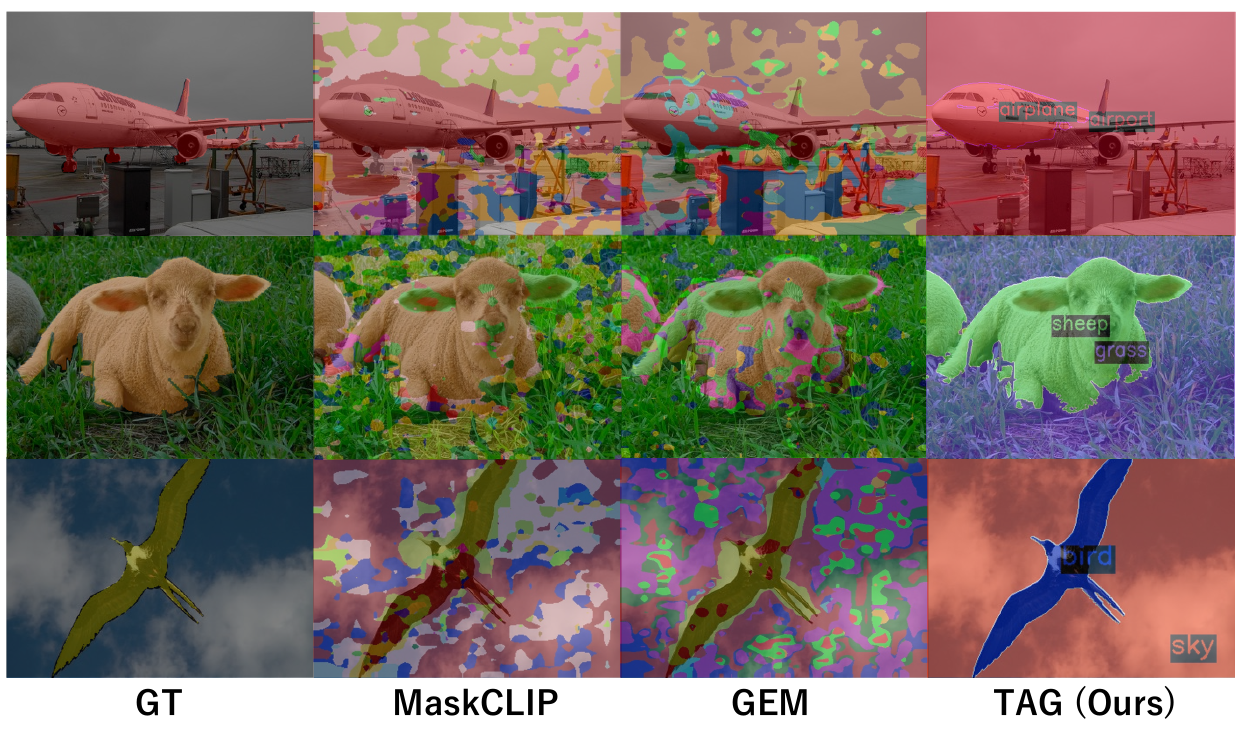}
    \caption{\textbf{Comparison results with CLIP base open-vocabulary segmentation methods on PascalVOC~\cite{voc}} Note that MaskCLIP~\cite{maskclip} and GEM~\cite{gem} uses text guidance while our TAG does not use.}
    \label{fig:clipbase}
\end{figure*}

\section{Experiment}
First, we present the implementation details in Section~\ref{sec:setting}.
Next, we compare our results to previous methods in Section~\ref{sec:main} and evaluate the open vocabulary aspect in Section~\ref{sec:web}. 
Finally, we justify the construction of TAG through an ablation study in Section~\ref{sec:ablation}.

\begin{figure*}[t]
    \centering
    \begin{minipage}[b]{\linewidth}
    \includegraphics[width=\linewidth]{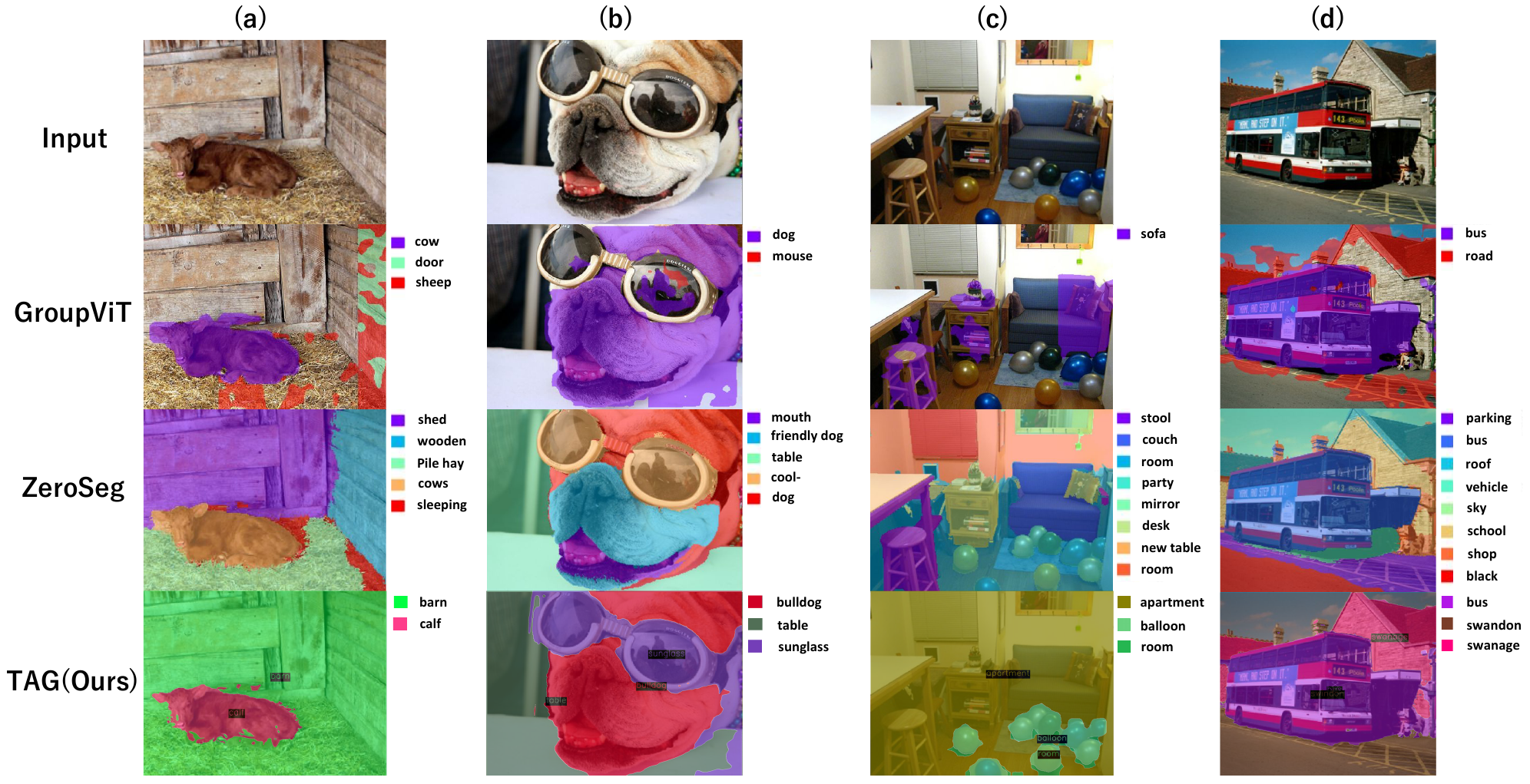}
    \caption{\textbf{Qualitative results.} We compare GroupViT~\cite{groupvit}, ZeroSeg~\cite{zeroseg}, and our TAG on images containing general objects from PascalContext~\cite{pc59}. This figure indicates that TAG can segment and label correctly.}
    \label{fig:main}
    \end{minipage}
\end{figure*}

\begin{table*}[t]
  \caption{\textbf{Comparison of state-of-the-art methods.} We evaluate on PascalVOC~\cite{voc}, PascalContext~\cite{pc59}, and ADE20K~\cite{ade} and report the mIoU. The clear boost in performance by retrieving open-vocabulary segment categories underlines their semantic richness and effectiveness.}
  \label{tab:main}
  \centering
  \scalebox{1.0}[1.0]{
  \begin{tabular}{l | c | c | c | c }
  \hline
    \multirow{2}{*}{Method} & ~Training~ & PascalVOC~\cite{voc} & PascalContext~\cite{pc59} & ADE20K~\cite{ade}\\
    \cline{3-5}
     & ~Free~ & 20class & 59class & 150class \\
     \hline
     \multicolumn{4}{l}{Open-Vocabulary Segmentation} \\
     \hline
     X-Decoder~\cite{xd} & - & 96.2 & 69.2 & 6.4 \\
     ODISE~\cite{odise} & - & 82.7 & 55.3  & 11.0 \\
     OpenSeg~\cite{openseg} & - & 72.2 & 44.8 & 8.8 \\
     GroupViT~\cite{groupvit} & - & 50.7 & 25.9 & - \\
     LSeg~\cite{lseg} & - & 47.4 & - & - \\
     GEM~\cite{gem} & $\surd$ & 26.5 & 11.8 & 4.4  \\
     MaskCLIP~\cite{maskclip} & $\surd$  & 28.6 & 25.5  & - \\
     \hline
     \multicolumn{4}{l}{Zero-Guidance Segmentation} \\
     \hline
     ZeroSeg~\cite{zeroseg} & $\surd$ & 20.1 & 19.6  & - \\
     SelfSeg~\cite{selfseg} & $\surd$ & 41.6 & -  & 6.4 \\
     TAG (Ours) & $\surd$ & \textbf{56.9} & \textbf{20.2}  & \textbf{6.6} \\
    \hline
  \end{tabular}
  }
\end{table*}

\subsection{Implementation Details}
\label{sec:setting}
For our implementation of TAG, we employed a frozen pre-trained CLIP~\cite{clip} and DINO~\cite{dinov2} with ViT-L/14 architecture and input $448 \times 448$ images to them.
As database, we use PMD~\cite{pmd}, CC12M~\cite{cc12m}, WordNet~\cite{wordnet} and EnglishWords~\cite{english}.
In addition, we use a fast indexing technique, FAISS~\cite{faiss}.
Our model works with a GPU memory of 15 GB.

\subsection{Main Results}
\label{sec:main}
To validate the performance of TAG, we conducted comprehensive comparative experiments with its closest counterpart, ZeroSeg~\cite{zeroseg}. 
For settings, TAG uses a PMD database~\cite{pmd}. We set the number of k-means clusters as 15 and the frequency filtering threshold 2.
For the evaluation, we used the mean Intersection over Union (mIoU) as the primary metric.
The predicted text $T_i$ needs to be assigned to one of the ground truth classes $T^{gt}$. $T_i$ is assigned to the ground-truth label that is closest in the Sentence-BERT~\cite{sbert} embedding space, following the same approach as ZeroSeg~\cite{zeroseg}.
Formally, the new label $T_i^*$ is computed by

\begin{align}
T_i^* = \underset{t \in T^{gt}}{\text{argmax}} \: [\text{cossim}^{\text{SBERT}}(T_i, t)] .
\end{align}

We perform our experiments on the PascalVOC~\cite{voc} dataset comprising 20 classes,  PascalContext~\cite{pc59} with 59 classes, as well as ADE20K~\cite{ade} consisting of 150 classes.

The qualitative results are shown in Figure~\ref{fig:clipbase} and Figure~\ref{fig:main}.
In Figure~\ref{fig:clipbase}, we compared TAG with CLIP~\cite{clip} base open-vocabulary method, MaskCLIP~\cite{maskclip}, and GEM~\cite{gem}. Using MaskCLIP~\cite{maskclip} and GEM~\cite{gem} results in a noisy and fragmented segmentation, whereas TAG achieves more consistent segments that better correspond with the shape of the object and segment categories.
In Figure~\ref{fig:main}, we compare GroupViT~\cite{groupvit}, ZeroSeg~\cite{zeroseg}, and our TAG on images containing general objects from PascalContext~\cite{pc59}.
In the image (a), TAG is the only method that accurately recognizes a cow as a calf. In addition, TAG assigns the precise and relevant class 'barn' to the surroundings of the calf, unlike ZeroSeg which incorrectly includes the class 'sleeping'.
In image (b), TAG is the only method to correctly identify 'sunglasses', and it also accurately classifies the dog as a 'bulldog'.
However, in image (c), TAG does not distinguish between a desk and a chair but rather assigns the rough class 'room' to the entire space. Occasionally, as shown in (d), TAG assigns proper nouns such as 'swindon' and 'swanage' which are names of cities in South England. While TAG correctly identifies the background as the city 'swanage', the ground is incorrectly assigned to the city of 'swindon'. We hypothesize this is caused by both segments being close in the CLIP~\cite{clip} embedding space.

The quantitative results are shown in Table~\ref{tab:main}.
TAG shows an improvement of +15.3 mIoU on PascalVOC~\cite{voc}, +0.6 mIoU on PascalContext~\cite{pc59}, and +0.2 mIoU on ADE20K~\cite{ade} compared to previous zero-guidance segmentation state-of-the-art results.
In particular, TAG shows a dramatic performance improvement on PascalVOC~\cite{voc}, which was identified as a limitation in ZeroSeg~\cite{zeroseg}.
Additionally, our TAG method has made significant improvements compared to untrained open-vocabulary segmentation methods, demonstrating an impressive improvement of +28.3 mIoU on PascalVOC~\cite{voc}, even without text-based guidance.

\begin{figure*}[t]
    \centering
    \begin{minipage}[b]{\linewidth}
    \includegraphics[width=\linewidth]{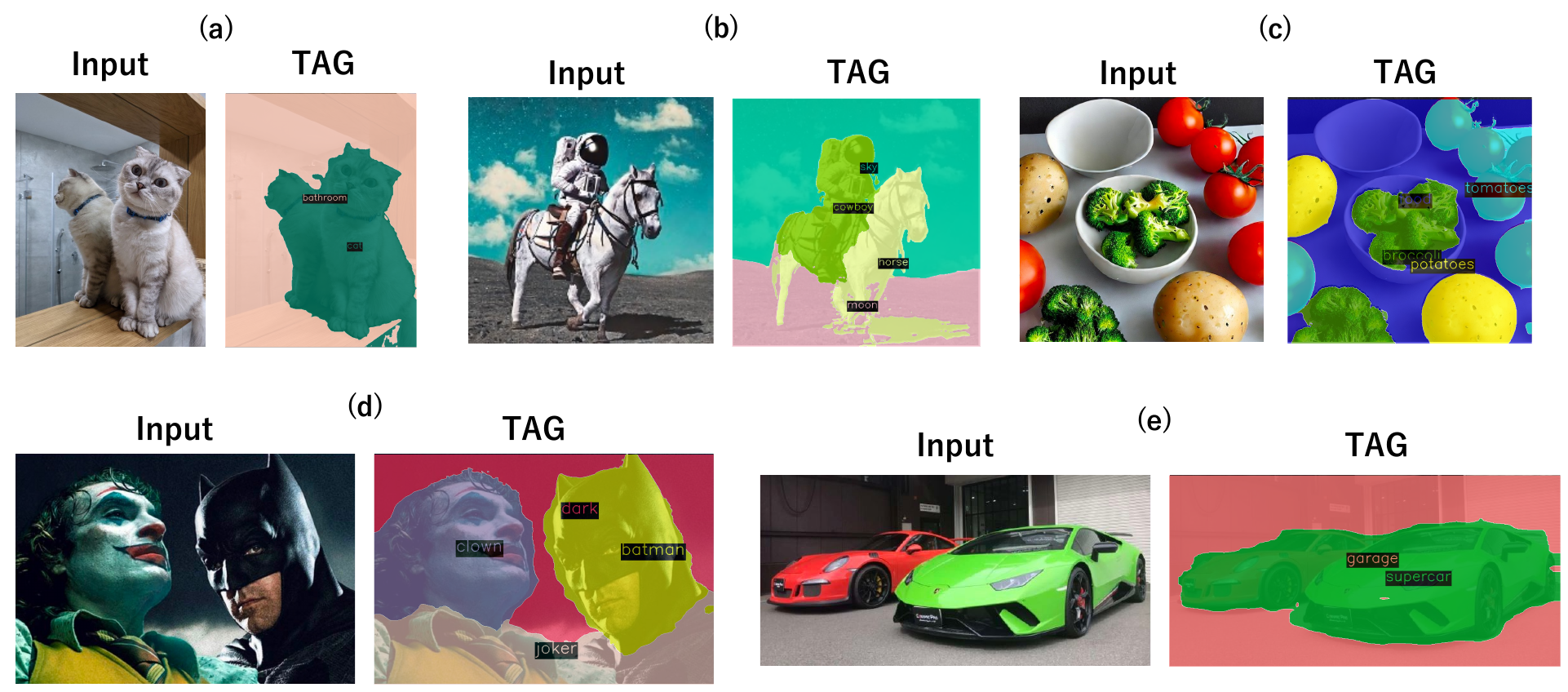}
    \caption{\textbf{Open-vocabulary segmentation results.} In (a) we test on a general image, (b) and (c) show images generated by Stable Diffusion~\cite{sd}, and (d) and (e) are images featuring specific proper nouns.}
    \label{fig:webimages}
    \end{minipage}
\end{figure*}

\subsection{Open Vocabulary Segmentation on Web-Crawled Images}
\label{sec:web}
In this section, we thoroughly assess the performance of TAG using open vocabulary segmentation experiments on web-crawled images, where we test the model's ability to accurately segment various unseen classes, including specific and detailed categories such as 'joker' and 'porsche'.

The qualitative outcomes of the experiments are visually depicted in Figure~\ref{fig:webimages}. In this figure, (a) represents a general image, while (b) and (c) showcase images created by Stable Diffusion~\cite{sd}. Furthermore, (d) and (e) show images containing proper nouns.

In image (a), although the complex concept of a 'mirror' is not captured, the segmentation successfully identifies both 'cat' and 'bathroom', resulting in accurate outcomes. In image (b), while failing to recognize the 'astronaut', the model aptly estimates the ground as the 'moon', leading to a logical result. Given TAG's ability to identify the ground as the moon, it is evident that it can understand the whole image while generating segment embeddings. Image (c) showcases the precise segmentation of various foods. Image (d) impressively segments and identifies proper nouns such as 'joker' and 'batman', demonstrating remarkable results. Lastly, image (e), despite containing the specific proper noun 'porsche', is correctly recognized as a supercar, affirming the accuracy of the segmentation.

These findings serve as compelling evidence that TAG exhibits robust capabilities to accurately segmenting open vocabularies, including complex and specific categories, thus underscoring its versatility and effectiveness in handling diverse and intricate segmentation tasks.

\begin{table*}[t]
  \caption{\textbf{Ablation study on database and numbers of cluster used for k-means.} We evaluate on PascalVOC~\cite{voc}, PascalContext~\cite{pc59}, and ADE20K~\cite{ade} and report the mIoU. These results reveal that using PMD~\cite{pmd} and CC12M~\cite{cc12m} as the database and setting the number of k-means clusters to 15 is the most robust choice, consistently yielding favorable outcomes across multiple datasets.}
  \label{tab:ablation}
  \centering
  \scalebox{1.0}[1.0]{
  \begin{tabular}{l | c | c | c | c }
  \hline
    \multirow{2}{*}{Database} & Cluster & PascalVOC~\cite{voc} & PascalContext~\cite{pc59} & ADE20K~\cite{ade}\\
    \cline{3-5}
     & Numbers & 20class & 59class & 150class \\
     \hline
     PMD~\cite{pmd} & 5 & 57.5 & 19.2 & 5.0 \\
     PMD~\cite{pmd} & 10 & 56.9 & 19.1 & 6.2 \\
     PMD~\cite{pmd} & 15 & 56.9 & \textbf{20.2} & \textbf{6.6} \\
     PMD~\cite{pmd} & 20 & 56.4 & 18.9 & 5.2 \\
     PMD~\cite{pmd} & 25 & 51.8 & 19.5 & 5.2 \\
     \hline
     CC12M~\cite{cc12m} & 5 & \textbf{58.5} & 19.2  & 4.9 \\
     CC12M~\cite{cc12m} & 10 & 58.4 & 19.2 & 6.1 \\
     CC12M~\cite{cc12m} & 15 & 58.1 & 19.6 & 6.1 \\
     CC12M~\cite{cc12m} & 20 & 57.6 & 19.0 & 5.0 \\
     CC12M~\cite{cc12m} & 25 & 52.9 & 19.1 & 5.1 \\
     \hline
     WordNet~\cite{wordnet} & 15 & 54.3 & 17.5 & 4.5 \\
     \hline
     EnglishWords~\cite{english} & 15 & 43.7 & 14.5 & 3.0 \\
    \hline
  \end{tabular}
  }
\end{table*}
\begin{figure*}[h]
    \centering
    \begin{minipage}[b]{\linewidth}
    \includegraphics[width=\linewidth]{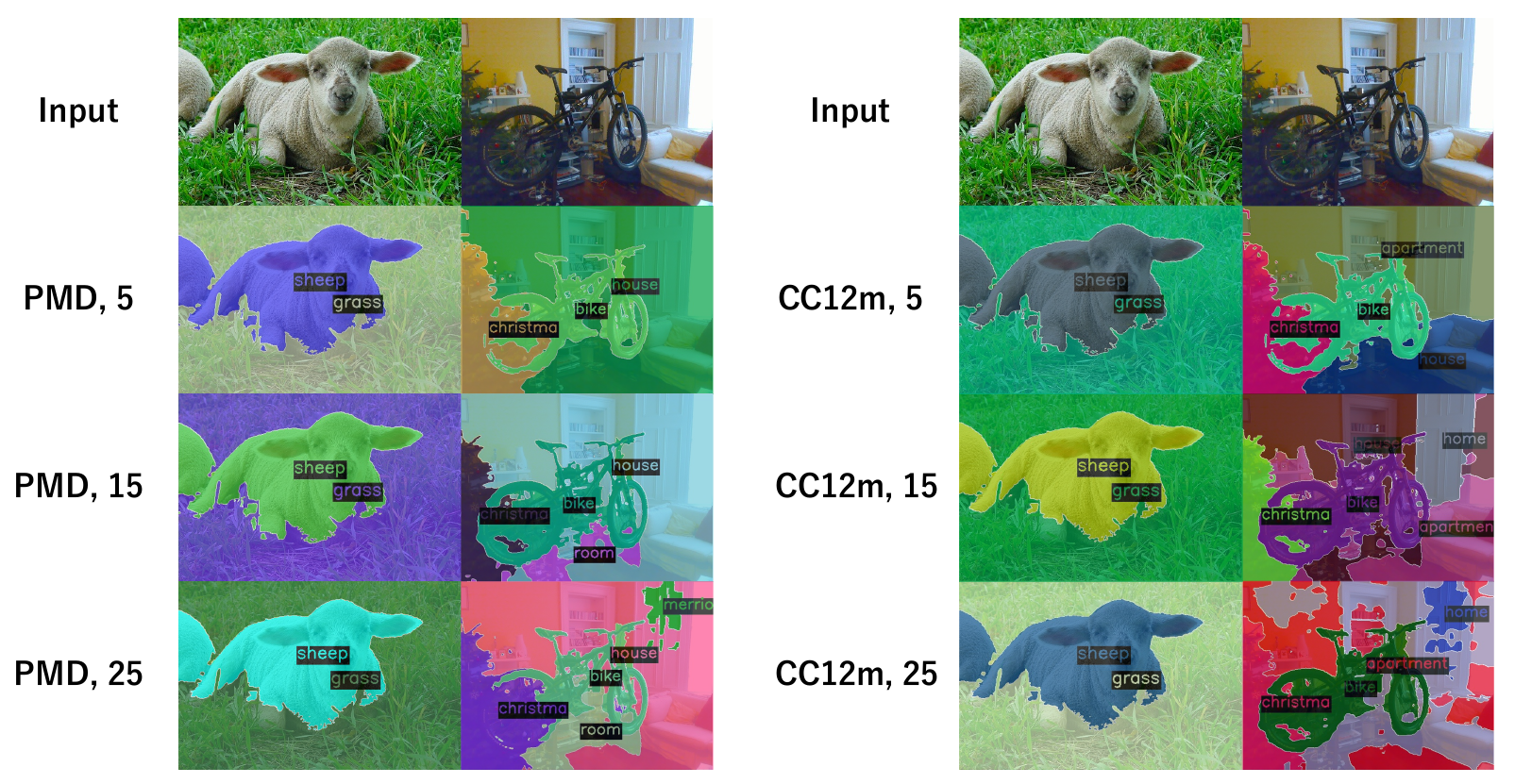}
    \caption{\textbf{Qualitative results of ablation study on PascalVOC~\cite{voc}.} The database and the number of k-means clusters are shown with the results.}
    \label{fig:ablation}
    \end{minipage}
\end{figure*}

\subsection{Ablation Study}
\label{sec:ablation}
In this section, we perform ablation studies on TAG, examining how various databases, cluster numbers of k-means, and label reassignment of evaluation affect performance.

Table~\ref{tab:ablation} presents the results of the ablation experiments comparing the effect of databases and cluster numbers of k-means on the mIoU.
The results indicate that PMD~\cite{pmd} and CC12M~\cite{cc12m} are preferable datasets for our database. These results also reveal that using PMD~\cite{pmd} as the database and setting the number of k-means clusters to 5 is the most robust choice, consistently yielding favorable outcomes across multiple datasets.
Figure~\ref{fig:ablation} shows the ablation qualitative results.
Left image remains unchanged regardless of variations in the database or the number of clusters, while increasing the number of clusters has been observed to cause segments with the same semantic meaning, such as 'apartment' to be divided into different segments like 'home' or 'house' in right image.

Furthermore, we conduct ablation experiments on filtering operations and the number of captions to select used for segment category retrieval.
Table~\ref{tab:ablation_filtering} shows that utilizing all three filtering operations yields the best results.
Similarly, we examine the effect of the threshold on the frequency filtering.
The results are shown in Table~\ref{tab:ablation_threshold} and indicate that using our default threshold of 2 is justified.
In addition, Table~\ref{tab:ablation_n} shows the number of captions to select and indicates that our default setting threshold to 10 is appropriate..

For evaluation, the predicted text $T_i$ is assigned to the ground-truth label that is closest in the Sentence-BERT~\cite{sbert} embedding space.
We conduct additional experiments on this assignment metrics. By calculating the IoU only for segments with a cosine similarity above a certain threshold, we enable the evaluation of TAG for different values of similarity. The experimental results are shown in Table~\ref{tab:ablation2}. The mIoU for segments with thresholds of 0.5 or higher demonstrates performance comparable to supervised, guided open-vocabulary segmentation methods as shown in Table~\ref{tab:ablation2}.
The experimental results also demonstrate that the inferred categories have at least a similarity score of zero or higher within the ground truth label, indicating that the predictions are not entirely off the mark.

\begin{table*}[t]
\vspace{3mm}
  \caption{\textbf{Ablation study on filtering operations.} We evaluate on PascalVOC~\cite{voc}, PascalContext~\cite{pc59}, and ADE20K~\cite{ade} and report the mIoU. This table shows that utilizing all three filtering operations yields the best results.}
  \label{tab:ablation_filtering}
  \centering
  \scalebox{1.0}[1.0]{
  \begin{tabular}{c  c  c | c | c | c }
  \hline
    \multirow{2}{*}{~Remove~} & \multirow{2}{*}{Standardize} & \multirow{2}{*}{~Filter~} & PascalVOC~\cite{voc} & PascalContext~\cite{pc59} & ADE20K~\cite{ade}\\
    \cline{4-6}
      & & & 20class & 59class & 150class \\
     \hline
      & & & 54.5 & 18.4 & 5.0 \\
      $\surd$ & & & 55.2 & 18.5 & 5.1 \\
      $\surd$ & $\surd$ &  & 56.0 & 18.6 & 5.8 \\
      $\surd$ & $\surd$ & $\surd$ & \textbf{56.9} & \textbf{20.2} & \textbf{6.6} \\

    \hline
  \end{tabular}
  }
\end{table*}



\begin{table*}[t]
  \caption{\textbf{Ablation study on the threshold of frequency filtering.} We evaluate on PascalVOC~\cite{voc}, PascalContext~\cite{pc59}, and ADE20K~\cite{ade} and report the mIoU.}
  \label{tab:ablation_threshold}
  \centering
  \scalebox{1.0}[1.0]{
  \begin{tabular}{c | c | c | c }
  \hline
    \multirow{2}{*}{threshold} & PascalVOC~\cite{voc} & PascalContext~\cite{pc59} & ADE20K~\cite{ade}\\
    \cline{2-4}
      & 20class & 59class & 150class \\
     \hline
     1  & 54.3 & 18.6 & 5.8 \\
     2 & \textbf{56.9} & \textbf{20.2} & 6.6 \\
     5  & 54.1 & 18.6 & \textbf{6.8} \\
     10  & 51.7 & 17.8 & 6.3 \\
     20 & 51.4 & 17.8 & 5.8 \\

    \hline
  \end{tabular}
  }
\end{table*}

\begin{table*}[t]
  \caption{\textbf{Ablation study on the number of retrieved captions.} We evaluate on PascalVOC~\cite{voc}, PascalContext~\cite{pc59}, and ADE20K~\cite{ade} and report the mIoU.}
  \label{tab:ablation_n}
  \centering
  \scalebox{1.0}[1.0]{
  \begin{tabular}{c | c | c | c }
  \hline
    Captions & PascalVOC~\cite{voc} & PascalContext~\cite{pc59} & ADE20K~\cite{ade}\\
    \cline{2-4}
    Number & 20class & 59class & 150class \\
     \hline
     5  & 55.9 & 19.5 & 6.3 \\
     10 & 56.9 & \textbf{20.2} & \textbf{6.6} \\
     20  & \textbf{57.3} & 19.9 & 6.3 \\
     40  & 56.4 & 19.2 & 6.0 \\
     60  & 55.6 & 19.1 & 5.9 \\

    \hline
  \end{tabular}
  }
\end{table*}

\begin{table*}[t]
\vspace{5mm}
  \caption{\textbf{Ablation study on the threshold of Sentence-BERT similarity.} We evaluate on PascalVOC~\cite{voc}, PascalContext~\cite{pc59}, and ADE20K~\cite{ade} and report the mIoU.}
  \vspace{-2mm}
  \label{tab:ablation2}
  \centering
  \scalebox{1.0}[1.0]{
  \begin{tabular}{c | c | c | c }
  \hline
    \multirow{2}{*}{threshold} & PascalVOC~\cite{voc} & PascalContext~\cite{pc59} & ADE20K~\cite{ade}\\
    \cline{2-4}
      & 20class & 59class & 150class \\
     \hline
     -1 & 56.9 & 20.2 & 6.6 \\
     0  & 56.9 & 20.2 & 6.6 \\
     0.5  & 79.5 & 24.4 & 8.9 \\
     0.8  & 84.8 & 32.8 & 15.0 \\

    \hline
  \end{tabular}
  }
\end{table*}

\section{Limitation}
While TAG achieves remarkable results, our proposed method still comes with certain limitations. First, as shown in Table~\ref{tab:ablation}, TAG depends on the choice of the database, making it challenging to select the optimal database for unknown domains without information on test labels. On the other hand, TAG can flexibly address this limitation by adding new concepts into the database without retraining, unlike language-based methods~\cite{zeroseg, selfseg}.
Second, TAG does not distinguish between different levels of class granularity. As shown in Figure~\ref{fig:webimages} (e), TAG predicted both 'Porsche' and 'Lamborghini' as 'supercar'. While the predicted categories in the qualitative results are consistently correct, they may not always align with the optimal category desired by the user. Future works might address this issue by considering the frequency of words within the database.

\section{Conclusion}
In this study, we proposed TAG, Training, Annotation, and Guidance-free open-vocabulary semantic segmentation. TAG employs a novel approach by extracting semantic features from each pixel in an image using CLIP~\cite{clip}, and then retrieving the open-vocabulary categories based on these features from an external database. Through a series of comprehensive experiments and analyses, we have demonstrated the effectiveness and versatility of TAG across various datasets and challenging segmentation tasks.
Our results indicate that TAG exhibits robust performance in handling diverse categories, including general classes and fine-grained, proper noun-based segments.

Overall, our findings highlight the potential of TAG as a powerful and effective tool in the field of semantic segmentation. By retrieving the open-vocabulary categories, we have successfully demonstrated the model's capability to handle diverse datasets and open vocabularies without text guidance, paving the way for future advancements and applications in this critical area of computer vision.

\clearpage  

%
%
\bibliographystyle{splncs04}
\bibliography{main}
\end{document}